\documentclass[10pt,twocolumn,letterpaper]{article}

\usepackage{cvpr}
\usepackage{enumitem}
\usepackage{times}
\usepackage{epsfig}
\usepackage{graphicx}
\usepackage{amsmath}
\usepackage{bm}
\usepackage{amssymb}
\usepackage{multirow}
\usepackage{algorithm}
\usepackage[noend]{algpseudocode}
\usepackage{makecell}
\usepackage{float}
\usepackage{subfig}
\usepackage{booktabs}
\usepackage{authblk}
\newcommand{\ra}[1]{\renewcommand{\arraystretch}{#1}}
\usepackage{soul}
\usepackage[font=small,labelfont=bf,tableposition=top]{caption}
\usepackage{etoolbox,lipsum}

\algrenewcommand\alglinenumber[1]{\footnotesize #1}

\usepackage[pagebackref=true,breaklinks=true,letterpaper=true,colorlinks,bookmarks=false]{hyperref}

\usepackage[breaklinks=true,bookmarks=false]{hyperref}

\cvprfinalcopy % *** Uncomment this line for the final submission

\def\cvprPaperID{****} % *** Enter the CVPR Paper ID here

\newcommand*\samethanks[1][\value{footnote}]{\footnotemark[#1]}
\author[1]{\vspace{-1.5em}Wuyang Chen\thanks{The first two authors contributed equally.}}
\author[1]{Ziyu Jiang\samethanks}
\author[1]{Zhangyang Wang}
\author[1]{Kexin Cui}
\author[2]{Xiaoning Qian\vspace{-0.8em}}
\affil[ ]{\textit {\{wuyang.chen, jiangziyu, atlaswang, ckx9411sx, xqian\}@tamu.edu}}
\affil[1]{Department of Computer Science and Engineering, Texas A\&M University}
\affil[2]{Department of Electrical and Computer Engineering, Texas A\&M University}
\affil[ ]{\small \url{https://github.com/chenwydj/ultra_high_resolution_segmentation}}

\begin{document}

%%%%%%%%% TITLE
\title{Collaborative Global-Local Networks for Memory-Efficient Segmentation of Ultra-High Resolution Images}

% \author{Wuyang Chen\thanks{The first two authors contributed equally.}, Ziyu Jiang\samethanks, Zhangyang Wang, Kexin Cui\\
% Department of Computer Science and Engineering\\
% Texas A\&M University\\
% {\tt\small \{wuyang.chen,jiangziyu,atlaswang,ckx9411sx\}@tamu.edu, xqian@ece.tamu.edu}
% \and
% Xiaoning Qian\\
% Department of Electrical and Computer Engineering\\
% Texas A\&M University\\
% {\tt\small xqian@ece.tamu.edu}
% }

\makeatletter
\let\@oldmaketitle\@maketitle% Store \@maketitle
\renewcommand{\@maketitle}{\@oldmaketitle
\vspace{-3.25em}
  \includegraphics[scale=0.45]{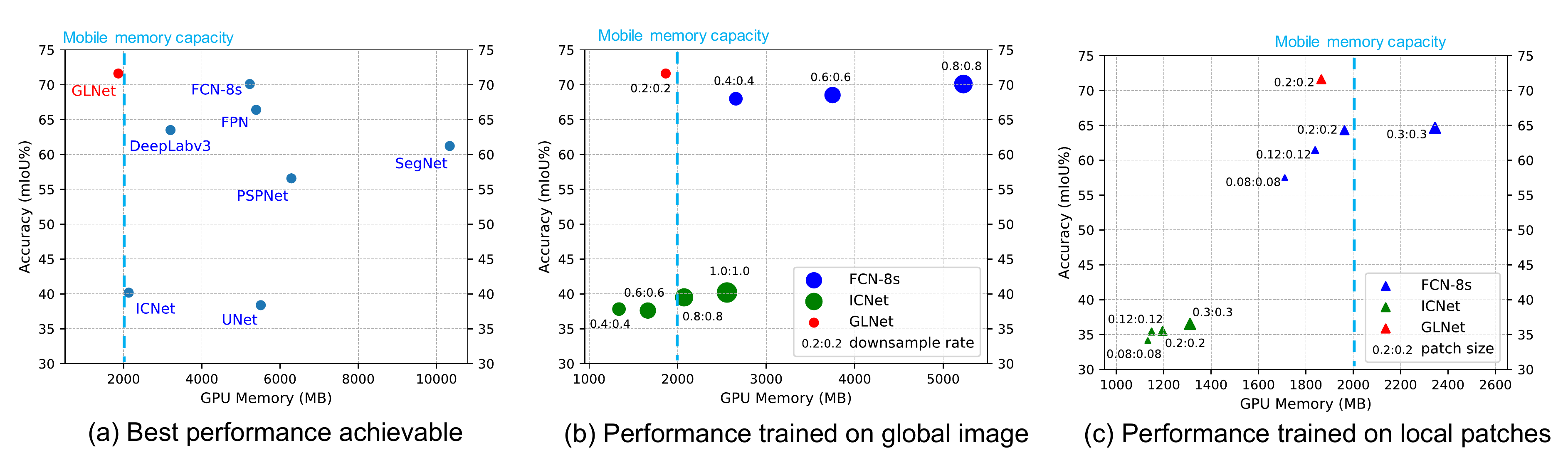}
  \vspace{-1.2em}
  \captionof{figure}{Inference memory and mean intersection over union (mIoU) accuracy on the DeepGlobe dataset \cite{demir2018deepglobe}. (a): Comparison of best achievable mIoU v.s. memory for different segmentation methods. (b): mIoU/memory with different global image sizes (downsampling rate shown in scale annotations). (c): mIoU/memory with different local patch sizes (normalized patch size shown in scale annotations). \textbf{GLNet} (red dots) integrates both global and local information in a compact way, contributing to a well-balanced trade-off between accuracy and memory usage. See Section \ref{sec:exp} for experiment details. Methods studied: ICNet \cite{zhao2018icnet}, DeepLabv3+ \cite{chen2018encoder}, FPN \cite{lin2017feature}, FCN-8s \cite{long2015fully}, UNet \cite{ronneberger2015u}, PSPNet \cite{zhao2017pyramid}, SegNet \cite{badrinarayanan2017segnet}, and the proposed GLNet.}
  \label{deep_globe_acc_mem}
  \bigskip
  }
\makeatother

\maketitle
%\thispagestyle{empty}

%%%%%%%%% ABSTRACT
\begin{abstract}
\vspace{-1em}
Segmentation of ultra-high resolution images is increasingly demanded, yet poses significant challenges for algorithm efficiency, in particular considering the (GPU) memory limits. Current approaches either downsample an ultra-high resolution image or crop it into small patches for separate processing. In either way, the loss of local fine details or global contextual information results in limited segmentation accuracy. We propose collaborative Global-Local Networks (GLNet) to effectively preserve both global and local information in a highly memory-efficient manner. GLNet is composed of a global branch and a local branch, taking the downsampled entire image and its cropped local patches as respective inputs. For segmentation, GLNet deeply fuses feature maps from two branches, capturing both the high-resolution fine structures from zoomed-in local patches and the contextual dependency from the downsampled input. To further resolve the potential class imbalance problem between background and foreground regions, we present a coarse-to-fine variant of GLNet, also being memory-efficient. Extensive experiments and analyses have been performed on three real-world ultra-high aerial and medical image datasets (resolution up to 30 million pixels). With only one single 1080Ti GPU and less than 2GB memory used, our GLNet yields high-quality segmentation results and achieves much more competitive accuracy-memory usage trade-offs compared to state-of-the-arts.
\vspace{-2em}
\end{abstract}

%Unlike previous works, we  met in some ultra-high resolution images

%%%%%%%%% BODY TEXT
\section{Introduction}

\begin{table*}
\centering
\captionsetup{font=normalsize}
\vspace{-1em}
\caption{Comparison of existing image segmentation datasets: the first three fall into the ultra-high resolution category.}
\label{dataset}
\resizebox{\textwidth}{!}{
\ra{1.0}
\begin{tabular}{@{}llllllcll@{}} \toprule
Dataset      &\phantom{abc}& Max Size               &\phantom{abc}& Average Size   &\phantom{abc}& \% of 4K UHR Images   &\phantom{abc}& \#Images \\ \midrule
DeepGlobe \cite{demir2018deepglobe}    && 6M pixels (2448$\times$2448)  && (uniform size) && 100\%                && 803      \\ 
ISIC \cite{tschandl2018ham10000, codella2018skin}        && 30M pixels (6748$\times$4499) && 9M pixels      && 64.1\%               && 2594     \\ 
Inria Aerial \cite{maggiori2017dataset} && 25M pixels (5000$\times$5000) && (uniform size) && 100\%                && 180      \\ \midrule

Cityscapes \cite{cordts2016cityscapes}   && 2M pixels (2048$\times$1024)  && (uniform size) && 0                    && 25000    \\ 
CamVid \cite{brostow2009semantic}       && 0.7M pixels (960$\times$720)  && (uniform size) && 0                    && 101      \\ 
COCO-Stuff \cite{caesar2018coco}    && 0.4M pixels (640$\times$640)  && 0.3M pixels    && 0                    && 123287   \\ 
VOC2012 \cite{everingham2010pascal}      && 0.25M pixels (500$\times$500) && 0.2M pixels    && 0                    && 2913     \\ \bottomrule
% ADE20K \cite{zhou2019semantic}      & 4M pixels (2100$\times$2100)  & 0.2M pixels    & ~ (only one image) & 22210    \\ \hline
\end{tabular}
}\vspace{-1em}
\end{table*}

%Histopathological Image

With the advancement of photography and sensor technologies, the accessibility to ultra-high resolution images has opened new horizons to the computer vision community and increased demands for effective analyses. Currently, an image with at least 2048$\times$1080 ($\sim$2.2M) pixels are regarded as \textit{2K high resolution media} \cite{ascher2007filmmaker}. An images with at least 3840$\times$1080 ($\sim$4.1M) pixels reaches the bare minimum bar of 4K resolution \cite{4k_resolution}, and \textit{4K ultra-high definition media} usually refers to a minimum resolution of 3840$\times$2160 ($\sim$8.3M) \cite{4k_resolution_full}.
Such images come %not only from next-generation television, but also 
from a wide range of scientific imaging applications, such as geospatial and histopathological images. Semantic segmentation allows better understanding and automatic annotations for these images. During the segmentation process, the image is pixel-wise parsed into different semantic categories, such as urban/forest/water areas in a satellite image, or lesion regions in a dermoscopic image. Segmentation of ultra-high resolution images plays important roles in a wide range of fields, such as urban planning and sensing \cite{volpi2017dense,robicquet2016learning}, as well as disease monitoring \cite{tschandl2018ham10000, codella2018skin}.

The recent development of deep convolutional neural networks (CNNs) has made remarkable progress in semantic segmentation. However, most models work on full resolution images and perform dense prediction, which requires more GPU memories comparing to image classification and object detection. This hurdle becomes significant when the image resolution grows to be ultra high, leading to the pressing dilemma between memory efficiency (even feasibility) and segmentation quality. Table \ref{dataset} lists a handful of existing ultra-high resolution segmentation datasets: DeepGlobe \cite{demir2018deepglobe}, ISIC \cite{tschandl2018ham10000, codella2018skin}, and Inria Aerial \cite{maggiori2017dataset}, in comparison to a few classical normal resolution segmentation datasets, to illustrate their drastic differences that result in new challenges. A more detailed discussion of the three ultra-high resolution datasets will be presented in Section \ref{sec:datasets}.

%\textbf{Status of Memory-Efficient Semantic Segmentation for Ultra-high Resolution Images:} 

Among the extensive research efforts on semantic segmentation, only limited attention have been devoted towards ultra-high resolution images.
%Specific memory-efficient segmentation approaches for ultra-high resolution images have rarely been studied. 
Typical ad-hoc strategies, such as downsampling or patch cropping, will result in the loss of either high-resolution details or spatial contextual information (see Section 3.1 for visual examples). Our in-depth studies show that high-accuracy methods like FCN-8s \cite{long2015fully} and SegNet \cite{badrinarayanan2017segnet} requires 5GB to 10GB of GPU memory to segment one 6M-pixel ultra-high resolution image during inference. These methods fall into the top-right area in Fig. \ref{deep_globe_acc_mem}(a) with high accuracy and high GPU memory usage. Contrarily, recent fast segmentation methods like ICNet \cite{zhao2018icnet}, whose memory usage is much alleviated, drops in its accuracy. These methods locate in the lower left corner in Fig. \ref{deep_globe_acc_mem}(a). Further studies with different sizes of global images and local patches (Fig. \ref{deep_globe_acc_mem} (b) and (c)) prove that typical models fail to achieve a good trade-off between the accuracy and the GPU memory usage.

\vspace{-0.5em}
\subsection{Our Contributions} 
\vspace{-0.5em}
This paper tackles memory-efficient segmentation of ultra-high resolution images, which presents \textbf{the first} dedicated analysis of this new topic to our best knowledge. The performance aim will be not only segmentation accuracy, but also reduced memory usage, and eventually, the trade-off between the two. 

\begin{figure}[t]
%\vspace{-0.1in}
\includegraphics[scale=0.35]{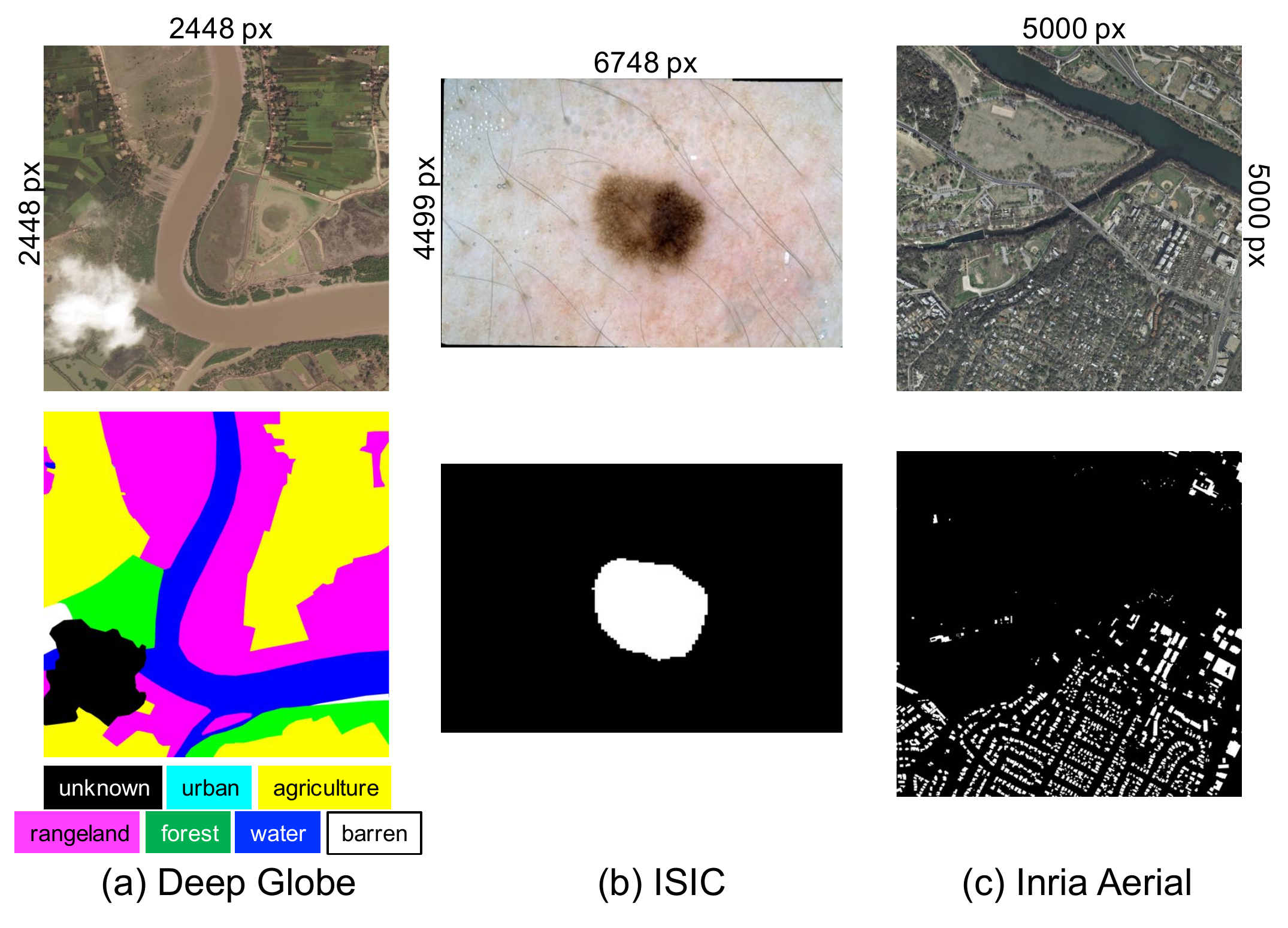}
\centering\vspace{-0.15in}
\caption{Three public datasets that fall into the ultra-high resolution category. DeepGlobe \cite{maggiori2017dataset} provides satellite images with 2448$\times$2448 pixels uniformly, labeled into seven categories of land regions. ISIC \cite{tschandl2018ham10000, codella2018skin} collects dermoscopy images of size up to 6748$\times$4499 pixels, with binary labels for segmenting foreground lesions. Inria Aerial \cite{maggiori2017dataset} provides binary masks for building/non-building areas in aerial images with  5000$\times$5000 pixels uniformly.}
\label{examples}\vspace{-1.3em}
\end{figure}

Our proposed model, named \textit{Collaborative Global-Local Networks} (\textbf{GLNet}), integrates both global images and local patches, for both training and inference. GLNet has a global branch and a local branch, handling downsampled global images and cropped local patches respectively. They further interact and ``modulate'' each other, through deeply shared and/or mutually regularized features maps across layers. This special design enables our GLNet the capability of well-balancing its accuracy and GPU memory usage (red dots in Fig. \ref{deep_globe_acc_mem}).
%During training and inference, the feature maps from low-level to high-level layers are deeply shared from one branch to the other. To enhance the collaboration between two branches, we also add feature map regularization and an aggregation layer for the final segmentation. 
To further resolve the class imbalance problem that often occurs, \textit{e.g.}, when one is primarily interested in segmenting small foreground regions, we provide a coarse-to-fine variant of our GLNet, where the global branch provides an additional bounding box localization. The GLNet design enables the seamless integration between global contextual information and necessary local fine details, balanced by learning, to ensure accurate segmentation. It meanwhile greatly trims down the GPU memory usage, as we only operate on downsampled global images plus cropped local patches; the original ultra-high resolution image is never loaded into the GPU memory. We summarize our main contributions as follows:
%\textcolor{red}{(ZW: Again this is your implementation, not the true innovation. I think you mean that you pass one patch a time, but you have to pass more patches as images get larger. Is that really invariant?}
%\st{We want to strictly control our memory usage and make it invariant to the input image size.} )
%This motivates us to leverage both downsampling and patch based training and inference. The GLNet has a global branch and a local branch, handling downsampled global image and cropped local patches respectively. During training and inference, the feature maps from low-level to high-level layers are deeply shared from one branch to the other. To enhance the collaboration between two branches, we also add feature map regularization and an aggregation layer for the final segmentation. To further on solve the class imbalance problem in ultra-high resolution images, we provide a coarse-to-fine version of our GLNet, where global branch provides an additional bounding box refinement. The whole design makes collaborative patch-based learning and inference well performed with a minimum requirement of GPU memory.
\vspace{-0.1in}
\begin{itemize}%[noitemsep,topsep=0pt]
    \item We develop a memory-efficient GLNet for the emerging new problem of ultra-high resolution image segmentation. The training requires only one 1080Ti GPU and inference requires less than 2GB GPU memory, for ultra-high resolution images of up to 30M pixels.\vspace{-0.1in}
    \item GLNet can effectively and efficiently integrate global context and local high-resolution fine structures, yielding high-quality segmentation. Either local or global information is proven to be indispensable.\vspace{-0.1in} %for our case
    \item We further propose a coarse-to-fine variant of GLNet to resolve the class imbalance problem in ultra-high resolution image segmentation, boosting the performance further while keeping the computation cost low.
    %\vspace{-0.1in}
\end{itemize}

%-------------------------------------------------------------------------
\section{Related Work}

\subsection{Semantic Segmentation: Quality \& Efficiency}
\vspace{-0.5em}
Fully convolutional network (FCN) \cite{long2015fully} was the first CNN architecture adopted for high-quality segmentation. U-Net \cite{ronneberger2015u,liu2018image,liu2018connecting} used skip-connections to concatenate low-level feature to high-level ones, with an encoder-decoder architecture. Similar structures were also adopted by DeconvNet \cite{noh2015learning} and SegNet \cite{badrinarayanan2017segnet}. DeepLab \cite{chen2014semantic, chen2018deeplab, yu2015multi, chen2018encoder} used dilated convolution to enlarge the field of view of filters. Conditional random fields (CRF) were also utilized to model the spatial relationship. Unfortunately, these models will suffer from prohibitively high GPU memory requirements when applied to ultra-high resolution images (Fig. \ref{deep_globe_acc_mem}).

As semantic segmentation grows important in many real-time/low-latency applications (\textit{e.g.} autonomous driving), efficient or fast segmentation models have recently gained more attention. ENet \cite{paszke2016enet} used an asymmetric encoder-decoder structure with early downsampling, to reduce the floating point operations. ICNet \cite{zhao2018icnet} cascaded feature maps from multi-resolution branches under proper label guidance, together with model compression. However, these models were not customized for nor evaluated on ultra-high resolution images, and our experiments show that they did not achieve sufficiently satisfactory trade-off in such cases.

%However, even the boost of speed and hardware utility, these models still lack specific ability for efficiently handling ultra-high resolution images, since the larger the image is, the worse the damage will be during the downsampling or compress process.

\subsection{Multi-Scale and Context Aggregation}
\vspace{-0.5em}
Multi-scale \cite{chen2014semantic, chen2016attention, xia2016zoom, hariharan2015hypercolumns} has proven to be powerful for segmentation, via integrating high-level and low-level features
%High level features contain more semantic meanings and less location information, and models with multi-scale ensembling are able to extract and 
to capture patterns of different granularity. In RefineNet \cite{lin2017refinenet}, a multi-path refinement block was utilized to combine multi-scale features via upsampling lower-resolution features. \cite{ghiasi2016laplacian} adopted a Laplacian pyramid to utilize higher-level features to refine boundaries reconstructed from lower-resolution maps. Feature Pyramid Networks (FPN) \cite{lin2017feature} progressively upsampled feature maps of different scales and aggregated them in a top-down fashion. Hierarchical Auto-Zoom Net (HAZN) \cite{xia2016zoom} utilized a two-step automatic zoom-in strategy to pass the coarse-stage bounding box and prediction scores to the finer stage.

Context aggregation also plays a key role in encoding the local spatial neighborhood, or even non-local information. Global pooling was adopted in ParseNet \cite{liu2015parsenet} to aggregate different levels of context for scene parsing. The dilated convolution and ASPP (atrous spatial pyramid pooling) module in DeepLab \cite{chen2018deeplab} helped enlarge the receptive field without losing feature map resolution too fast, leading to the aggregation of global contexts into local information. Similar goal was accomplished by the pyramid pooling in PSPNet \cite{zhao2017pyramid}. In ContextNet \cite{poudel2018contextnet}, BiSeNet \cite{yu2018bisenet} and GUN \cite{mazzini2018guided}, the deep/shallow branches were combined to aggregate global context and high-resolution details. \cite{visin2016reseg} considered the contextual information as a long-range dependency modeled by RNNs.
% It is worth noting that multi-scale and context aggregation can be in both input and feature map levels. In GLNet, both input (global/local branch) and feature map level aggregations are adopted.
It is worth noting that in our GLNet, the context aggregation is adopted in both input level (global/local branch) and feature level.

\subsection{Ultra-high Resolution Segmentation Datasets} \label{sec:datasets}
\vspace{-0.5em}
%It is both a challenge and a precious opportunity (\textcolor{red}{keng}) that the 

%So far, ultra-high resolution datasets are much fewer than standard segmentation datasets. 
%less well-studied and quantitatively fewer than the datasets for middle to high resolution images. 
We summarize three public datasets with ultra-high images (studied in Section \ref{sec:exp}). Basic information and visual examples are shown in Table \ref{dataset} and Fig. \ref{examples}, respectively.

The DeepGlobe Land Cover Classification dataset (\textbf{DeepGlobe}) \cite{demir2018deepglobe} is the first public benchmark offering high-resolution sub-meter satellite imagery focusing on rural areas. DeepGlobe provides ground truth pixel-wise masks of seven classes: urban, agriculture, rangeland, forest, water, barren, and unknown. It contains 1146 annotated satellite images, all of size 2448$\times$2448 pixels. DeepGlobe is of significantly higher resolution and more challenging than previous land cover classification datasets. 

The International Skin Imaging Collaboration (\textbf{ISIC}) \cite{tschandl2018ham10000, codella2018skin} dataset collects a large number of dermoscopy images. Its subset, the ISIC Lesion Boundary Segmentation dataset, consists of 2594 images from patient samples presented for skin cancer screening. All images are annotated with ground truth binary masks, indicating the locations of the primary skin lesion.
%a large-scale publicly accessible dataset of dermoscopy images. Currently, their dataset houses images from leading clinical centers internationally, acquired from a variety of devices used at each center. The ISIC Lesion Boundary Segmentation Challenge provides lesion segmentation from dermoscopic images in the form of binary masks. 
Over 64\% images have ultra-high resolutions: the largest image has 6682$\times$4401 pixels.

The \textbf{Inria Aerial} Dataset \cite{maggiori2017dataset} covers diverse urban landscapes, ranging from dense metropolitan districts to alpine resorts. It provides 180 images (from five cities) of 5000$\times$5000 pixels, each annotated with a binary mask for building/non-building areas. Different from DeepGlobe, it splits the training/test sets by city instead of random tiles.

% Their coverage is 810 km$^2$ and has aerial orthorectified color imagery with a spatial resolution of 0.3 m. The images cover dissimilar urban settlements, ranging from densely populated areas (e.g. San Francisco's financial district) to alpine towns (e.g. Lienz in Austrian Tyrol).

%-------------------------------------------------------------------------
\section{Collaborative Global-Local Networks}

\subsection{Motivation: Why Not Global or Local Alone}
\begin{figure}[ht]
\includegraphics[scale=0.2]{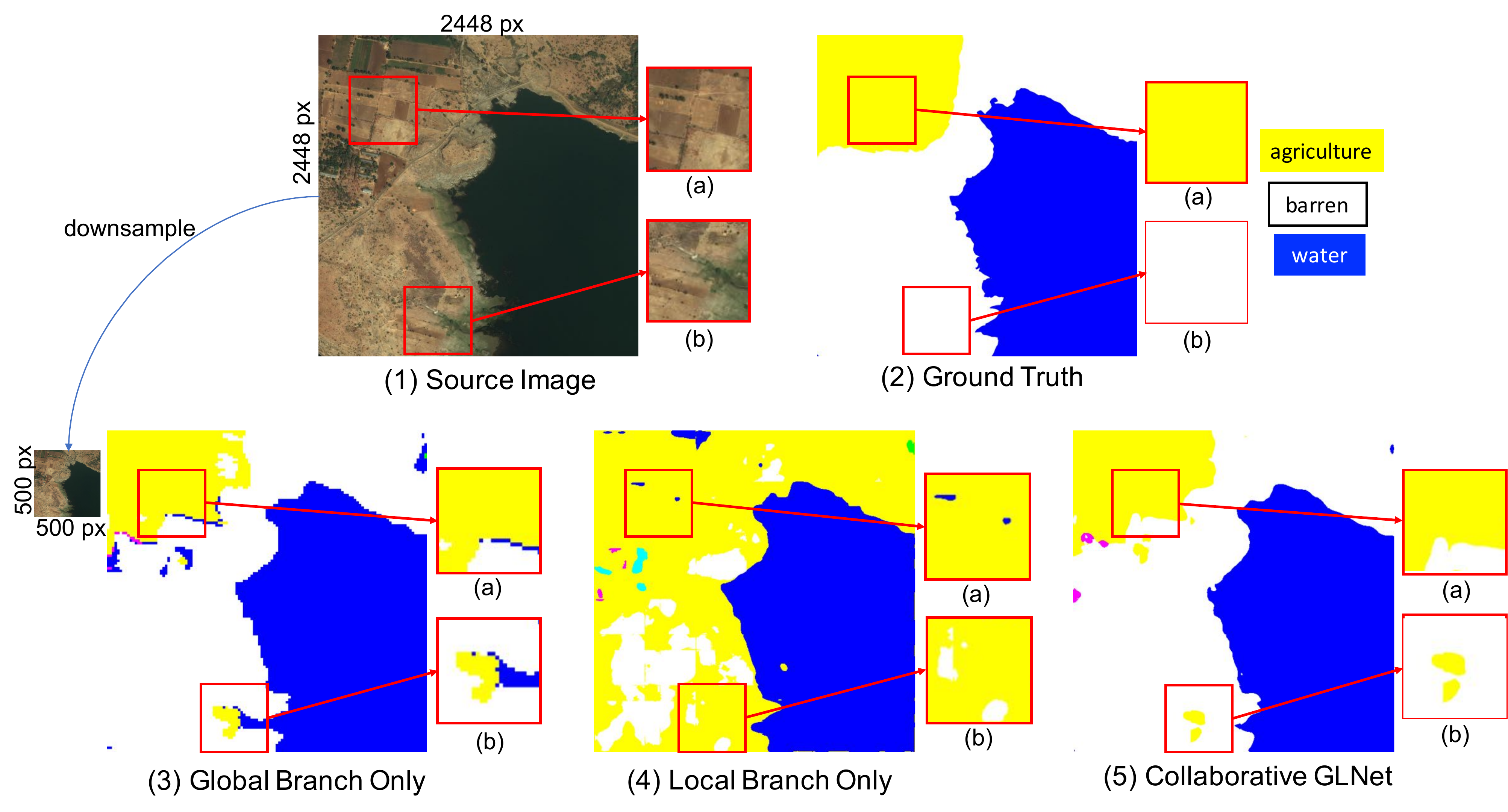}
\centering \vspace{-0.1in}
\caption{Example segmentation results in DeepGlobe dataset (best viewed in a high-resolution display): (1) Source image. (2) Ground-truth segmentation mask. We show predictions by (3) model trained with downsampled global images only, (4) model trained with cropped local patches only, (5) our proposed collaborative GLNet. The zoom-in panels (a) and (b) illustrate the details of local fine structures, showing the undesired grid-like artifacts and inaccurate boundaries from the global or local result alone.}
\label{deep_globe_example}\vspace{-0.12in}
\end{figure}

\vspace{-0.5em}
For training and inference on ultra-high resolution images with limited GPU memory, two ad-hoc ideas may come up first: downsampling the global image, or cropping it into patches. However, they both often lead to undesired artifacts and poor performance. 
Fig. \ref{deep_globe_example}(1) displays a 2448$\times$2448-pixel image, with its ground-truth segmentation in Fig. \ref{deep_globe_example}(2): yellow represents ``agriculture'', blue  ``water'', and white ``barren''.  We then trained two FPN models: one with all images downsampled to 500$\times$500 pixels, the other with cropped patches of size 500$\times$500 pixels from original images. Their predictions are displayed in Fig. \ref{deep_globe_example}(3) and (4), respectively. One can observe that the former suffers from ``jiggling'' artifacts and inaccurate boundaries, due to the missing details from downsampling. In comparison, the latter has large areas misclassified. Note that ``agriculture'' and ``barren'' regions often visually look similar (zoom-in panels (a) and (b) in Fig. \ref{deep_globe_example}(1)). Therefore, the patch-based training lacks spatial contexts and neighborhood dependency information, making it difficult to distinguish between ``agriculture'' and ``barren'' using local patches only. Finally, we provide our GLNet' prediction in Fig. \ref{deep_globe_example}(5) for reference: it clearly shows the advantages of leverage merits from both global and local processing.

%Using downsampled global image of size 500$\times$500 (leftmost in Fig. \ref{deep_globe_example}), the prediction (Fig. \ref{deep_globe_example}(3)) segments the region roughly correct, but with grid-like artifacts and inaccurate boundaries. This undesired prediction owes to the loss of details during downsampling. If one switch to patch-based training (with 500$\times$500 patch size, same as the downsampled global image), during inference the model will fail to distinguish similar patches, due to the lack of large receptive field and long-term dependencies (Fig. \ref{deep_globe_example}(4)).

%Small images are required for memory-efficient training and inference. The failure of the downsampled global image and local patches alone makes us believe that to achieve both high quality and high efficiency in segmentation for ultra-high resolution image, we must leverage merits from both global image and local patches. We propose our deep feature fusion strategy in Section 3.2.

\subsection{GLNet Architecture}

%The main motivation behind our architecture design is to minimize GPU memory usage during both training and inference while preserving segmentation accuracy. We adopt patch-based learning and inference throughout our model since it is a memory-invariant approach to handle an image with arbitrarily large size. In this section, we focus our main effort on collaborating both global context and local details.

%\textcolor{red}{Still, ``We adopt patch-based learning and inference throughout our model since it'' might evoke critiques.} \textcolor{blue}{why?}

\begin{figure*}
\includegraphics[scale=0.44]{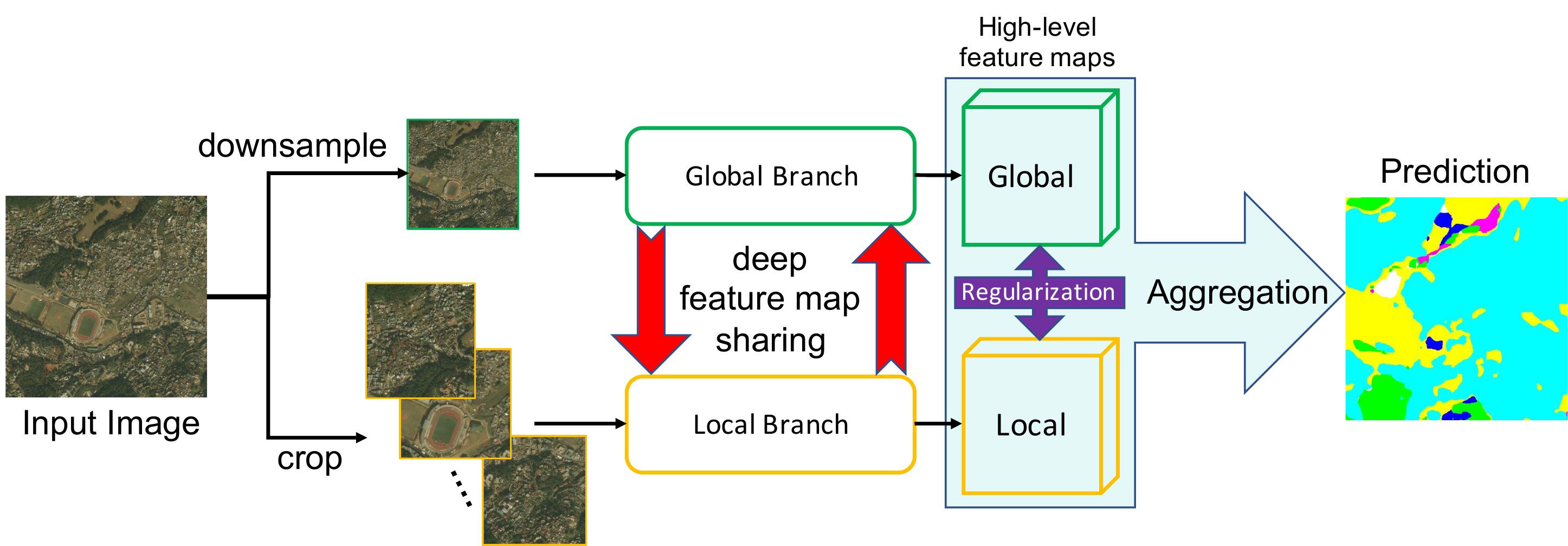}
\centering 
%\vspace{-0.1in}
\caption{Overview of our proposed GLNet. The global and local branch takes downsampled and cropped images, respectively. Deep feature map sharing and feature map regularization enforce our global-local collaboration. The final segmentation is generated by aggregating high-level feature maps from two branches.}\vspace{-0.12in}
\label{glnet}
\end{figure*}

\subsubsection{The Global and Local Branches}

We depict our GLNet architecture in Fig. \ref{glnet}. Starting from the dataset of $N$ ultra-high resolution images and segmentations $\mathcal{D} = \{(\boldsymbol{I}_i, \boldsymbol{S}_i)\}_{i=1}^N$ where $\boldsymbol{I}_i, \boldsymbol{S}_i \in \mathbb{R}^{H \times W}$, the global branch
$\mathcal{G}$
% $\mathcal{M}^{\text{Glb}}$
takes down-sampled low-resolution images $\mathcal{D}^{\text{lr}} = \{(\boldsymbol{I}_i^{\text{lr}}, \boldsymbol{S}_i^{\text{lr}})\}_{i=1}^N$, and the local branch
$\mathcal{L}$
% $\mathcal{M}^{\text{Loc}}$ 
receives cropped patches from 
% the source image
$\mathcal{D}$
with the same resolution $\mathcal{D}^{\text{hr}} = \{\{(\boldsymbol{I}_{ij}^{\text{hr}}, \boldsymbol{S}_{ij}^{\text{hr}})\}_{j=1}^{n_i}\}_{i=1}^N$, where each $\boldsymbol{I}_i$ and $\boldsymbol{S}_i$ in $\mathcal{D}$ comprises $n_i$ patches. Note that $\boldsymbol{I}_i$ and $\boldsymbol{S}_i$ were fully cropped into patches (instead of random cropping) to facilitate both training and inference.
$\boldsymbol{I}_i^{\text{lr}}, \boldsymbol{S}_i^{\text{lr}} \in \mathbb{R}^{h_1 \times w_1}$ and $\boldsymbol{I}_i^{\text{hr}}, \boldsymbol{S}_i^{\text{hr}} \in \mathbb{R}^{h_2 \times w_2}$, where $h_1, h_2 \ll H$, and $w_1, w_2 \ll W$.
% The output will be the segmentation map for the current input local patch, with the help of contextual information provided by the (downsampled) global image.
We adopt the same backbone for $\mathcal{G}$ and $\mathcal{L}$, both can be viewed as a cascade of convolutional blocks from layer $1$ to $L$ (Fig. \ref{gl_branch}).

%We leverage our feature map deep sharing strategy (Section 3.2.2) for the global-local collaboration. 
During the segmentation process, the feature maps from all layers of either branch are deeply shared with the others (Section \ref{sec:deep_fm_share}). Two sets of high-level feature maps are then aggregated to generate the final segmentation mask via a branch aggregation layer $f_{\text{agg}}$ (Section \ref{sec:agg_reg}). To constrain the two branches and stabilize training, a weakly-coupled regularization is also applied to the local branch training.\vspace{-0.1in}

\subsubsection{Deep Feature Map Sharing} \label{sec:deep_fm_share}

%Our feature sharing strategy act as follows: 
To collaborate with the local branch, feature maps from the global branch are first cropped at the same spatial location of the current local patch and then upsampled to match the size of the feature maps from the local branch. Next, they are concatenated as extra channels to the local branch feature maps in the same layer. In a symmetrical fashion, the feature maps from the local branch are also collected. The local feature maps are first downsampled to match the same relative spatial ratio as the patches were cropped from the large source image. Then they are merged together (in the same order as the local patches were cropped) into a complete feature map of the same size as the global branch feature map. Those local feature maps are also concatenated as channels to the global branch feature maps, before feeding into the next layer.

\begin{figure}[t]
\includegraphics[scale=0.28]{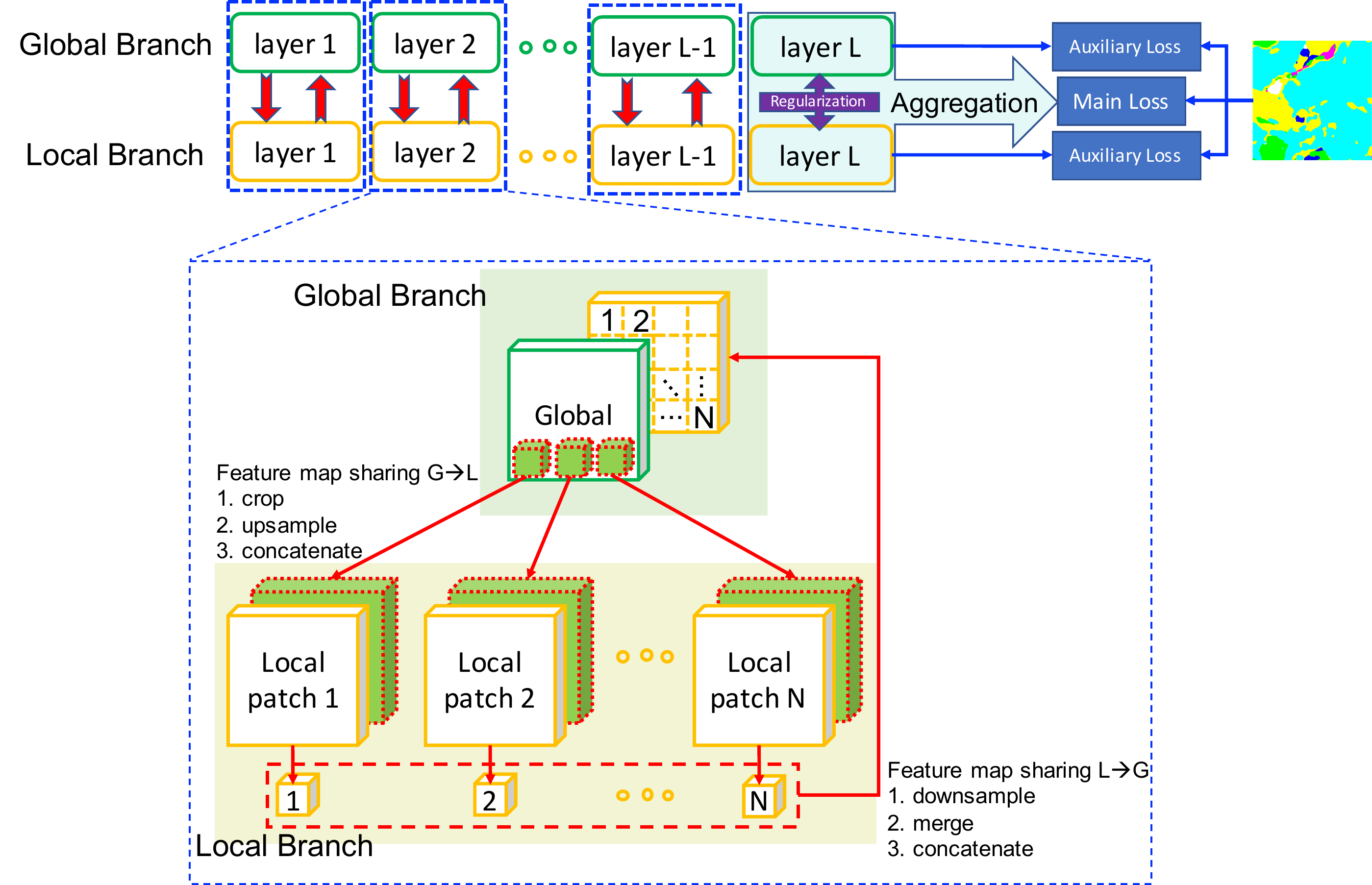}
\centering 
%\vspace{-0.1in}
\caption{Deep feature map sharing between the global and local branch. At each layer, feature maps with global context and ones with local fine structures are bidirectionally brought together, contributing to a complete patch-based deep global-local collaboration. The main loss from the aggregated results and two auxiliary losses from two branches form our optimization target.} 
\vspace{-0.12in}
\label{gl_branch}
\end{figure}

Fig. \ref{gl_branch} illustrates the process of deep feature map sharing, which is applied layer-wise except the last layer of branches. The sharing direction can be either unidirectional (\textit{e.g.} sharing global branch's feature maps to local branch, $\mathcal{G}\rightarrow\mathcal{L}$) or bidirectional ($\mathcal{G}\rightleftarrows\mathcal{L}$). At each layer, the current global contextual features and local fine structural features take reference and are fused to each other.\vspace{-0.1in}

%\textcolor{red}{Since we use concatenation for the feature map sharing, the channel size is increased during the fusion. Therefore in our GLNet, we use a basic global branch $\mathcal{M}^{\text{Glb}_0}$ as a starter to provide feature maps for our local branch. This basic global branch does not consume any external feature maps. (ZW: I totally don't understand this.)} \textcolor{blue}{Since we use channel-wise concatenation for the feature map sharing, the channel sizes of our convolution filters in each branch are doubled after the fusion. Therefore in our GLNet, we use a basic global branch $\mathcal{M}^{\text{Glb}_0}$ as a starter to provide feature maps for our local branch. This basic global branch does not consume any external feature maps.}

%To thoroughly facilitate the collaboration between two branches and utilize difference scales of feature maps, in our GLNet this feature sharing strategy is adopted in stages $1, 2, ..., L-1$, except for the final high-level layer $L$. At each stage, feature maps with global context and ones with local fine structures are always brought together, contributing to a complete patch-based deep global-local collaboration, without sacrificing any extra computing resources. We show our complete deep feature map sharing strategy in Fig. \ref{gl_branch}.

\subsubsection{Branch Aggregation with Regularization} \label{sec:agg_reg}

The two branches will be aggregated
% to make final prediction
through an aggregation layer $f_{\text{agg}}$, implemented as a convolutional layer of 3$\times$3 filters.
% , whose channel number equals to the number of segmentation classes. 
It takes the high-level feature maps from the local branch's $L^{\text{th}}$ layer $\boldsymbol{\hat{X}}_{L}^{\text{Loc}}$, and same ones from the global branch $\boldsymbol{\hat{X}}_{L}^{\text{Glb}}$, and concatenate them along the channel. The output of $f_{\text{agg}}$ will be the final segmentation output $\boldsymbol{\hat{S}}^{\text{Agg}}$. In addition to the main segmentation loss enforced on $\boldsymbol{\hat{S}}^{\text{Agg}}$, we also apply two auxiliary losses, 
% with the same weight as the main loss, 
to enforce the segmentation output from the local branch $\boldsymbol{\hat{S}}^{\text{Loc}}$ and from the global branch $\boldsymbol{\hat{S}}^{\text{Glb}}$ to be close to their corresponding segmentation maps (local patch / global downsampled), respectively, which we find helpful for stabilizing the training.% Specifically, we use the Focal Loss \cite{lin2018focal} as our loss function.

%In our experiments, we proved that the aggregation prediction out-performs each individual prediction, $\boldsymbol{\hat{S}}^{\text{Glb}}$ from the global branch and $\boldsymbol{\hat{S}}^{\text{Loc}}$ from the local branch. Two auxiliary losses for the global and local branch are also applied in addition to the main loss from the aggregation result.

We find in practice that the local branch is prone to overfitting some strong local details, and ``overriding'' the learning of global branch. Therefore, we try to avoid the local branch from learning ``too much faster'' than the global one, by adding a weakly-coupled regularization between feature maps from the last layers of two branches.
%Since local patches provide limited receptive field and lack global context, it is still possible that the prediction of the local branch over-fits local patterns without considering the global context. To avoid this problem, for the last high-level layer in our GLNet, instead of using the feature sharing method mentioned above, we use feature map regularization between the global branch and local branch as our fusion strategy.
Specifically, we add the Euclidean norm penalty $\lambda \|\boldsymbol{\hat{X}}_{L}^{\text{Loc}} - \boldsymbol{\hat{X}}_{L}^{\text{Glb}}\|_2$ to discourage large relative changes between $\boldsymbol{\hat{X}}_{L}^{\text{Loc}}$ and $\boldsymbol{\hat{X}}_{L}^{\text{Glb}}$ with $\lambda$ empirically fixed as 0.15 in our work. This regularization is mainly designed to make local branch training ``slow down'' and more synchronized with global branch learning, and it only updates the parameters in local branch.
% we ``turn off'' this regularization by switching $\lambda$ to zero, when training the global branch. 

%at each training step

%Note that we only apply this L2 constraint when training the local branch. We experimentally found that the L2 norm did not perform well when training the global branch.

\subsection{Coarse-to-Fine GLNet}
\vspace{-0.5em}
%\subsection{Class Imbalance in Ultra-high Resolution Segmentation}

% \begin{figure}[ht]
% \includegraphics[scale=0.31]{images/isic_fb_pixel_ratio_crop_example.pdf}
% \centering
% \caption{Left: Histogram of the ratio of foreground pixels to background pixels (per image) in ISIC 2018 Training Set. Right: Class-imbalanced examples from ISIC dataset.}
% \label{isic_example}
% \end{figure}

%A popular case in than. However, many  have fairly small and concentrated foreground, causing , in the ISIC training set
For segmentation to separate foreground and background (\ie, binary masks), the foreground
% usually contains objects of interests, which 
often takes little space in ultra-high resolution images. Such class imbalance may seriously damage the segmentation performance. Taking the ISIC dataset for example, $\sim$99\% of images have more background than foreground pixels, and over 60\% of images have less than 20\% foreground pixels (see the blue bars in Fig. \ref{isic_bbox_hist}(1)). Many local patches will contain nothing but background pixels, which leads to ill-conditioned gradients.\vspace{-0.15in}

%This is a very serious class imbalance problem. In this scenario, if one uses downsampled global images for training, each mini-batch will contain gradients from both foreground pixels and background pixels. However, the patch-based learning will make things worse, since it is very likely that all image patches in a mini-batch contain only background pixels which back-propagates highly biased gradients. We show an example from the ISIC dataset in Fig. \ref{isic_example} right panel. This example image has a size of 4401$\times$6682 (29,407,482 pixels), and the area of foreground (5,566,107 pixels) is much smaller than the background (5566107 / 29407482 $\approx$ 18.9\%).

% for both training and inference
\paragraph{A Two-Stage Refinement Solution}

\begin{figure}[!ht]
\includegraphics[scale=0.27]{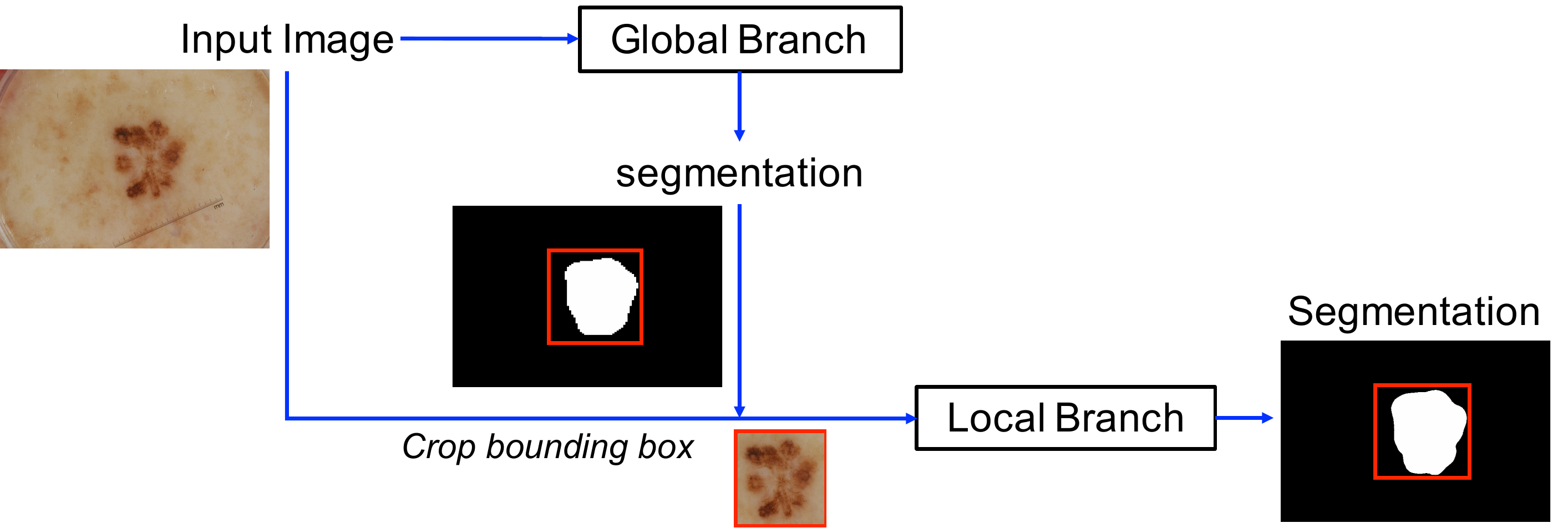}
\centering
\caption{Two-stage segmentation. Our global branch does coarse segmentation, and for fine segmentation with the local branch we only process the bounding box foreground-centered region.}
%\vspace{-1em}
\label{two_step_seg}
\end{figure}

To alleviate the class imbalance, we propose a novel two-stage coarse-to-fine variant of GLNet (Fig. \ref{two_step_seg}). It first applies the global branch alone to fulfill a coarse segmentation on downsampled images. A bounding box is then created for the segmented foreground region\footnote{In practice, we dynamically relax the bounding box size, so that the bounded region has a foreground-background class ratio around 1, to have class balance for the second step.}. The bounded foreground in the original full-resolution image is then fed as the input for the local branch for fine segmentation. Different from GLNet admitting parallel local-global branches, this \textit{Coarse-to-Fine GLNet} admits a sequential composition of the two branches, where the feature maps only within the bounding box are first deeply shared from the global to the local branch during the bounding box refinement, and then shared back. All regions beyond the bounding box will be predicted as the background. The Coarse-to-Fine GLNet also reduces the computation cost, through selective fine-scale processing.

\section{Experiments} \label{sec:exp}
%the following experiments

In this section, we evaluate the performance of GLNet on the DeepGlobe and Inria Aerial datasets and evaluate the efficacy of the coarse-to-fine GLNet on the ISIC dataset. We thoroughly compare our models with other methods to show both the segmentation quality and memory efficiency\footnote{We have chosen several state-of-the-art models with public implementations for comparison (see supplementary for more elaborations)}. Ablation study is also carefully presented. 

\subsection{Implementation Details}

In our work, we adopt the FPN (Feature Pyramid Network) \cite{lin2017feature} with ResNet50 \cite{he2016deep} as our backbone. The deep feature map sharing strategy is applied on the feature maps from conv2 to conv5 blocks of ResNet50 in the bottom-up stage, and also on the feature maps from the top-down and smoothing stages in the FPN. For the final lateral connection stage in the FPN, we adopted the feature map regularization, and aggregate this stage for the final segmentation. For simplicity, both the downsampled global image and the cropped local patches share the same size, 500$\times$500 pixels. Neighboring patches have 50-pixel overlap to avoid boundary vanishing for all the convolutional layers. We use the Focal Loss \cite{lin2018focal} with $\gamma = 6$ as the optimization target for both the main and two auxiliary losses. Equal weights (1.0) are assigned to the main and auxiliary losses. The feature map regularization coefficient $\lambda$ is set to 0.15.

To measure the GPU memory usage of a model, we use the command line tool ``gpustat'', with the minibatch size of 1 and avoid calculating any gradients. Note that only a single GPU card is used for our training and inference. 

We conduct experiments using the PyTorch framework \cite{paszke2017automatic}. We use the Adam optimizer \cite{kingma2014adam} ($\beta_1 = 0.9$, $\beta_2 = 0.999$) with learning rate of $1\times10^{-4}$ for training the global branch, and $2\times10^{-5}$ for the local branch. We use a minibatch size of 6 for all training. All experiments are performed on a workstation with NVIDIA 1080Ti GPU cards.% under CUDA 9.0 environment.

% \subsection{Training Strategy}
% Note that to fully facilitate the collaboration between two branches, we always fully crop the whole image into patches, not randomly cropping.
% Our training starts with the basic global branch using downsampled global images. Once the basic global branch converges, we freeze this branch and start the local branch training together with the aggregation layer. Finally, we freeze the local branch and train the global branch and fine-tune the aggregation layer.

\subsection{DeepGlobe}

We first apply our framework to the DeepGlobe dataset. This dataset contains 803 ultra-high resolution images (2448$\times$2448 pixels). We randomly split images into training, validation and testing sets with 455, 207, and 142 images respectively. The dense annotation contains 7 classes of landscape regions, where one class out of seven called ``unknown'' region is not considered in the challenge.

\begin{table*}[h]

\caption{Efficacy of different feature map sharing strategies evaluated on the local DeepGlobe test set. `Agg' stands for the aggregation layer and `Fmreg' means feature map Euclidean norm regularization. `$\mathcal{G}\rightarrow\mathcal{L}$' and `$\mathcal{G}\rightleftarrows\mathcal{L}$' represent feature map sharing from the global to local branch and bidirectionally between two branches respectively. `Shallow' and `deep' denote whether sharing feature maps in a single layer or in all layers in a model.}
{\footnotesize
\centering
    \label{deep_globe_ablation_results}
    \ra{1.0}
    \resizebox{\textwidth}{!}{
    \begin{tabular}{@{}lccccccccccccc@{}} \toprule
    \multirow{2}{*}{Model} &\phantom{ab} & \multirow{2}{*}{Agg} &\phantom{abc}& \multirow{2}{*}{Fmreg} &\phantom{abc} & \multicolumn{2}{c}{$\mathcal{G}\rightarrow\mathcal{L}$} &\phantom{abc}& $\mathcal{G}\rightleftarrows\mathcal{L}$ &\phantom{abc}& \multirow{2}{*}{mIoU (\%)} &\phantom{abc}& \multirow{2}{*}{Memory (MB)} \\ \cmidrule{7-8} \cmidrule{10-10}
     &&&&&& shallow  & deep && deep & & \\ \midrule
    Local only &&&  &  &&  &&  &&  & 57.3 && 1189 \\ 
    Global only &&&  &  &&  &&  &&  & 66.4 && 1189 \\ \midrule
    \multirow{5}{*}{GLNet} && \checkmark &&  &&  &&  &&  & 69.3 && 1189 \\ 
     && \checkmark && \checkmark &&  &  &&  && 70.3 && 1209 \\  
     && \checkmark && \checkmark && \checkmark &  &&  && 70.5 && 1251 \\ 
     && \checkmark && \checkmark &&  & \checkmark &&  && 70.9 && 1395 \\
     && \checkmark && \checkmark &&  & \checkmark &&\checkmark && \textbf{71.6} && 1865 \\ \bottomrule
    \end{tabular}
    }
}
\end{table*}

\subsubsection{From shallow to deep feature map sharing}

To evaluate the performance of our global-local collaboration strategy, we progressively upgrade our model from shallow to deep feature map sharing (Table \ref{deep_globe_ablation_results}). With downsampled global images or image patches alone, each branch could only achieve the mean intersection over union (mIoU) of 57.3\% and 66.4\% respectively. By the aggregation of the high-level feature maps from two branches and the regularization between them, the performance can be boosted to 70.3\%. When we share only a single layer of feature maps from the global to the local branch (``shallow sharing''), the aggregated results increased by 0.2\%, and when we upgrade to ``deep sharing'' where feature maps of all layers are shared, the mIoU is rocked to 70.9\%. Finally, the bidirectional deep feature map sharing between two branches enables the model to yield a high mIoU of 71.6\%.

This ablation study proves that, with deep and diverse feature map sharing/regularization/aggregation strategies, the global and local branch can effectively collaborate together. It is worth noting that even with the bidirectional deep feature map sharing approach (last row in Table \ref{deep_globe_ablation_results}), the memory usage during inference is only slightly increased from 1189MB to 1865MB.

\begin{figure}[ht]
\includegraphics[scale=0.15]{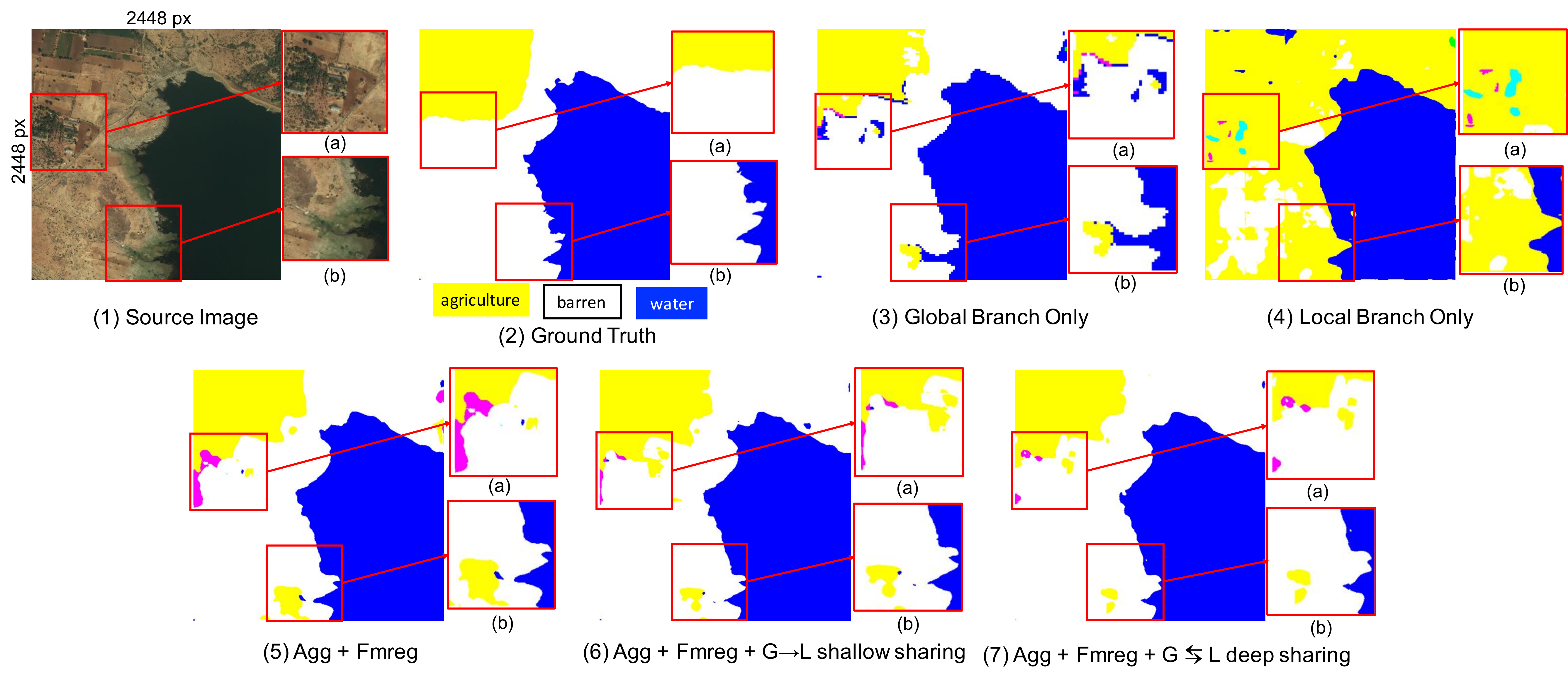}
\centering\vspace{-0.1in}
\caption{Example segmentation results in DeepGlobe dataset (best viewed in a high-resolution display). (1) Source image. (2) Ground truth. We show predictions by models trained with: (3) downsampled global images only, (4) cropped local patches only, (5) aggregation (`Agg') and feature map regularization (`Fmreg'), (6) shallow feature map sharing, and (7) bidirectional deep feature map sharing. Improvements from (3) to (7) in the zoom-in panels (a) and (b) illustrate the details of local fine structures.}
\label{deep_globe_ablation_example}
\end{figure}

Fig. \ref{deep_globe_ablation_example} visualizes the achieved improvements with two zoom-in panels (a) and (b) showing details. There are undesired grid-like artifacts and inaccurate boundaries in the global (Fig. \ref{deep_globe_ablation_example}(3)) or local results (Fig. \ref{deep_globe_ablation_example}(4)) alone. From aggregation, shallow feature map sharing, and finally to bidirectional deep feature map sharing, progressive improvements can be observed with both significantly reduced misclassification and inaccurate boundaries.

%are dramatically removed.

\subsubsection{Accuracy and memory usage comparison\protect\footnote{we use public available segmentation models \cite{mshahsemseg, kazutodeeplab}}}

Models trained and inferenced with global images or local patches may yield different results. This is because models have different receptive fields, convolution kernel sizes, and padding strategies, which results in different suitable training/inference choices. Therefore we carefully compared models trained with these two approaches in this ablation study. We train and test a model twice (with global images or local patches each time), and then pick its best result.

\paragraph{Coarse comparison with fixed image/patch size\protect\footnote{Since some models (\textit{e.g.} SegNet, PSPNet) cannot process images without downsampling during global inference due to heavy memory usage, they have to be trained with downsampled global images. We avoid over-downsampling to reduce the loss of resolution. For patch-based training and inference, we adopted 500$\times$500 pixels for all models.}} \hspace{1cm} 
Table \ref{deep_globe_final_results} shows that all models achieve higher mIoU under global inference, but consume very high GPU memories. Their memory usages drop in patch-based inference, but accuracies also nose dive. Only our GLNet achieves the best trade-off between mIoU and GPU memory usage. We plot the best achievable mIoU of each method in Fig. \ref{deep_globe_acc_mem}(a).

\begin{table}[!h]
\caption{Predicted mIoU and inference memory usage on the local DeepGlobe test set. `$\mathcal{G}\rightarrow\mathcal{L}$' and `$\mathcal{G}\rightleftarrows\mathcal{L}$' means feature map sharing from the global to local branch and bidirectionally between two branches respectively. Note that our GLNet does not inference with global images. See Fig. \ref{deep_globe_acc_mem}(a) for visualization.}
{\small
\centering
\label{deep_globe_final_results}
\ra{1.0}
\resizebox{\columnwidth}{!}{
\begin{tabular}{@{}lcccccc@{}}\toprule
\multirow{2}{*}{Model}& \multicolumn{2}{c}{Patch Inference} & \phantom{a} & \multicolumn{2}{c}{Global Inference} \\ \cmidrule{2-3} \cmidrule{5-6}
 & mIoU(\%) & Memory(MB) && mIoU(\%) & Memory(MB) \\ \midrule
UNet\cite{ronneberger2015u} & 37.3 & 949 && 38.4 & 5507 \\
ICNet\cite{zhao2018icnet} & 35.5 & 1195 && 40.2 & 2557 \\ 
PSPNet\cite{zhao2017pyramid} & 53.3 & 1513 && 56.6 & 6289 \\ 
SegNet\cite{badrinarayanan2017segnet} & 60.8 & 1139 && 61.2 & 10339 \\ 
DeepLabv3+\cite{chen2018encoder} & 63.1 & 1279 && 63.5 & 3199 \\ 
FCN-8s\cite{long2015fully} & 64.3 & 1963 && 70.1 & 5227 \\ \midrule
& \multicolumn{2}{c}{mIoU(\%)} & \phantom{a} & \multicolumn{2}{c}{Memory(MB)} \\\midrule
\textbf{GLNet: $\mathcal{G}\rightarrow\mathcal{L}$} & \multicolumn{2}{c}{\textbf{70.9}} & \phantom{a} & \multicolumn{2}{c}{\textbf{1395}} \\ 
\textbf{GLNet: $\mathcal{G}\rightleftarrows\mathcal{L}$} & \multicolumn{2}{c}{\textbf{71.6}} & \phantom{a} & \multicolumn{2}{c}{\textbf{1865}}  \\ \bottomrule
\end{tabular}
}
}
\end{table}

\paragraph{In-depth comparison with different image/patch sizes}
We select FCN-8s and ICNet for in-depth evaluations with different image/patch sizes, since they achieve high mIoU and efficient memory usage respectively. We plot details of this ablation study in Fig. \ref{deep_globe_acc_mem}(b) and (c). For both FCN-8s and ICNet, higher accuracy means to sacrifice GPU memory usage, and vice versa. This proves that typical models fail to balance their segmentation quality and efficiency\footnote{In training with large global images, the minibatch size is limited by the heavy memory usage. We adopt the ``late update'' optimization trick, \textit{e.g.} a minibatch size of 2 with weights updated every three minibatches.}.

\subsection{ISIC\protect\footnote{The ISIC Lesion Boundary Segmentation challenge uses the following metrics (per image): score = 0 if IoU $<$ 0.65; score = IoU, otherwise.}}

The ISIC Lesion Boundary Segmentation Challenge dataset contains 2594 ultra-high resolution images. We randomly split images into training, validation and testing sets with 2074, 260, and 260 images respectively.% As the foreground-background ratios are highly biased, we apply our proposed coarse-to-fine segmentation. % for this dataset.

\subsubsection{Coarse-to-fine segmentation}

\begin{figure}[ht]
\begin{center}
    \includegraphics[scale=0.42]{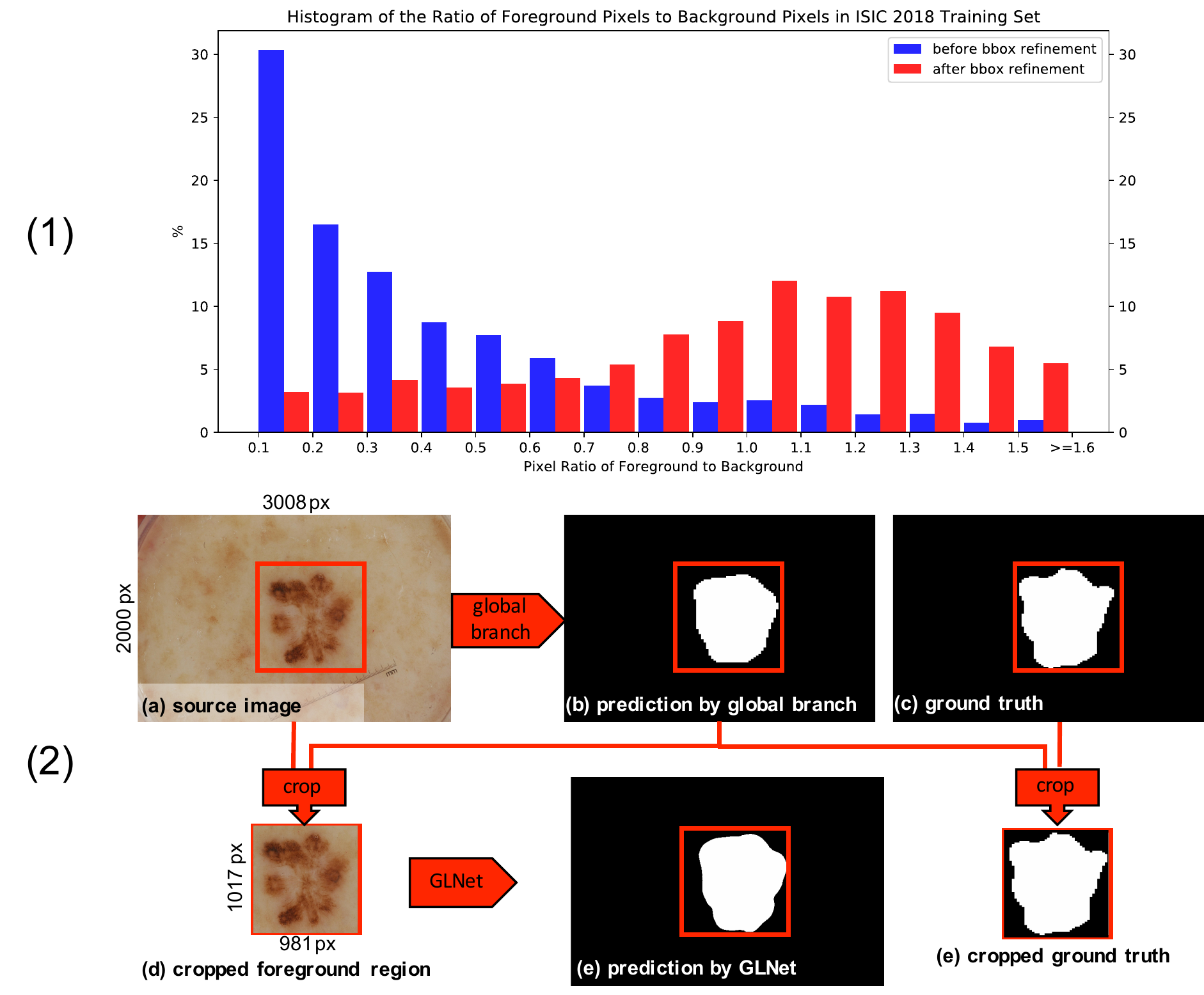}
    \caption{(1) Histogram of the ratio of foreground to background pixels (per image) in ISIC 2018 dataset. Blue bars represent ratios before global branch's bounding box refinement, and red for ratios after the refinement, which is more balanced. (2) Visual results of coarse-to-fine segmentation. After the refinement (from (b) to (e)) the GLNet is able to capture more accurate boundaries.}
    \label{isic_bbox_hist}
\end{center}
\end{figure}

On the heavily imbalanced ISIC dataset, the global and local branch can only achieve 72.7\% and 48.5\% mIoU respectively. When we apply our coarse-to-fine strategy, we can clearly see a much more balanced foreground-background class ratio (red bars in Fig. \ref{isic_bbox_hist}(1)).

By cropping a relaxed bounding box for the foreground (Section 4.2), the local branch is only trained on smaller and class-balanced images, and the cropped-out margin region is assumed as background by default. With class-balanced images, the global-to-local sharing strategy yields 73.9\% mIoU, and further bidirectional sharing boosts the performance to 75.2\%. In this scenario, the global branch uses a more accurate global context since there is less information loss during downsampling the cropped smaller images. This success proves that the coarse-to-fine segmentation can better capture the context information and solve the class-imbalance problem. We list the results of this ablation study in Table \ref{isic_ablation_results} and some visual results in Fig. \ref{isic_bbox_hist}(2).

\begin{table}[!htb]
\caption{Efficacy of coarse-to-fine segmentation and deep feature map sharing evaluated on local ISIC test set. `$\mathcal{G}\rightarrow\mathcal{L}$' and `$\mathcal{G}\rightleftarrows\mathcal{L}$' represent feature map sharing direction from the global to local branch and bidirectionally between two branches respectively. `Bbox' means bounding box refinement by the global branch.}
{\tiny
\centering
\label{isic_ablation_results}
\ra{1.0}
\resizebox{\columnwidth}{!}{
\begin{tabular}{@{}lccccc@{}}\toprule
Model &\phantom{a}& $\mathcal{G}\rightarrow\mathcal{L}$ & $\mathcal{G}\rightleftarrows\mathcal{L}$ & Bbox & mIoU (\%) \\ \midrule
Global only &&  &  &  & 70.1 \\
Local only &&  &  &  & 48.5 \\ \midrule
\multirow{3}{*}{GLNet} && \checkmark &  &  & 72.7 \\  
 && \checkmark &  & \checkmark & 73.9 \\
 &&  & \checkmark & \checkmark & \textbf{75.2} \\ \bottomrule
\end{tabular}
}
}
\end{table}

\subsubsection{Accuracy and memory usage comparison\protect\footnote{Since images in ISIC are class-imbalanced, training with downsampled global images is the best strategy for most of the methods. Therefore, for each method we choose a proper image size to balance the information loss during downsampling and the GPU memory usage.}}

We finally list mIoU and inference memory usage of GLNet on the local test set of the ISIC dataset in Table \ref{isic_final_results}. GLNet yields mIoU 75.2\% and is quantitatively better than other methods on both accuracy and memory usage.

\begin{table}[!h]
\caption{Predicted mIoU and inference memory usage on the local ISIC test set.}
{\scriptsize
\centering
\label{isic_final_results}
\resizebox{\columnwidth}{!}{
\ra{1.0}
\begin{tabular}{@{}lcccc@{}} \toprule
Model &\phantom{abcd}& mIoU (\%) &\phantom{abcd}& Memory (MB) \\ \midrule
% UNet && 57.1 && 5507 \\ 
ICNet\cite{zhao2018icnet} && 33.8 && 1593 \\ 
% PSPNet && 72.2 && 1721 \\
SegNet\cite{badrinarayanan2017segnet} && 37.1 && 4213 \\ 
DeepLabv3+\cite{chen2018encoder} && 70.5 && 2033 \\ 
FCN-8s\cite{long2015fully} && 42.8 && 5238 \\ 
% FPN &&  && 5389 \\ 
\textbf{GLNet} && \textbf{75.2} && \textbf{1921}\\ \bottomrule
\end{tabular}
}
}
\end{table}

\subsection{Inria Aerial}
% \subsubsection{Accuracy and memory usage comparison}

The Inria Aerial Challenge dataset contains 180 ultra-high resolution images, each with 5000$\times$5000 pixels. We randomly split images into training, validation and testing sets with 126, 27, and 27 images respectively. Table \ref{aerial_ablation_results} demonstrates the efficacy and efficiency of our deep feature map sharing strategy. The testing results are listed in Table \ref{aerial_final_results}, where our proposed GLNet yields mIoU 71.2\%. Again, GLNet is quantitatively better than other methods on both accuracy and memory usage. It is worth noting that our GLNet preserves a low memory usage even for a ``super'' ultra-high resolution image with 5000$\times$5000 pixels.

\begin{table}[!htb]
\caption{Efficacy of the GLNet evaluated on local Inria Aerial test set. `$\mathcal{G}\rightarrow\mathcal{L}$' and `$\mathcal{G}\rightleftarrows\mathcal{L}$' represent feature map sharing direction from the global to local branch and bidirectionally between two branches respectively.}
{\footnotesize
\centering
\label{aerial_ablation_results}
\ra{1.0}
\resizebox{\columnwidth}{!}{
\begin{tabular}{@{}lcccccc@{}} \toprule
Model & \phantom{abc} & $\mathcal{G}\rightarrow\mathcal{L}$ & \phantom{abc} & $\mathcal{G}\rightleftarrows\mathcal{L}$ & \phantom{abc} & mIoU (\%) \\ \midrule
Global only &&  &&  && 42.5 \\ 
Local only &&  &&  && 63.1 \\ \midrule
\multirow{2}{*}{GLNet} && \checkmark &&  && 66.0 \\  
  & & & & \checkmark && \textbf{71.2} \\ \bottomrule
\end{tabular}
}
}
\end{table}

\begin{table}[!htb]
\caption{Predicted mIoU and inference memory usage on local Inria Aerial test set.}
{\small
\centering
\label{aerial_final_results}
\ra{1.0}
\resizebox{\columnwidth}{!}{
\begin{tabular}{@{}lcccc@{}} \toprule
Model & \phantom{abcdefg} & mIoU (\%) & \phantom{abcdefg} & Memory (MB) \\ \midrule
% UNet &  & 5507 \\ \hline
ICNet\cite{zhao2018icnet} && 31.1 && 2379 \\ 
% PSPNet &  & 6289 \\ \hline
% SegNet &  & 10339 \\ \hline
DeepLabv3+\cite{chen2018encoder} && 55.9 && 4323 \\ 
FCN-8s\cite{long2015fully} && 61.6 && 8253 \\ 
% FPN &  & 5389 \\ \hline
\textbf{GLNet} && \textbf{71.2} && \textbf{2663} \\ \bottomrule
\end{tabular}
}
}
\end{table}

\section{Conclusions}

We proposed a memory-efficient segmentation model GLNet specifically for the ultra-high resolution images. It leverages both the global context and local fine structure effectively to enhance the segmentation in the scenario of ultra-high resolution without sacrificing the GPU memory usage. We also proved that the class imbalance problem can be solved by our coarse-to-fine segmentation approach.

We believe that pursuing an optimal balance of GPU memory and accuracy is essential for the study of ultra-high resolution images, which makes our model important. Our work is pioneering this new research topic of memory-efficient segmentation of the ultra-high resolution images.

\section*{Acknowledgement}
The work of Z. Wang is in part supported by the National Science Foundation Award RI-1755701. The work of X.Qian is in part supported by the National Science Foundation Award CCF-1553281. We also thank Prof. Andrew Jiang and Junru Wu for helping experiments.

{\small
\bibliographystyle{unsrt}
\bibliography{egbib}
}

\end{document}